%% file: ms.tex
\documentclass[12pt]{article}

\usepackage{times}
\usepackage{soul}
\usepackage{url}
\usepackage[small]{caption}
\usepackage{graphicx} 
\usepackage{bmpsize}
\usepackage{amsmath}
\usepackage{booktabs}
\usepackage{algorithm}
\usepackage{algorithmic}
\usepackage[dvipsnames]{xcolor}
\usepackage{multirow}
\usepackage{siunitx}

\title{Using theoretical ROC curves for analysing machine learning binary classifiers}

\author{
\parbox{1.0\textwidth}
{\centering
Luma Omar and Ioannis Ivrissimtzis\\
Durham University\\
Department of Computer Science\\
Durham, DH1 3LE, UK\\
\{luma.omar, ioannis.ivrissimtzis\}@durham.ac.uk
}
}

\begin{document}	

\date{}
\maketitle
		
\begin{abstract}
Most binary classifiers work by processing the input to produce a scalar response and comparing it to a threshold value. The various measures of classifier performance assume, explicitly or implicitly, probability distributions $P_s$  and $P_n$ of the response belonging to either class, probability distributions for the cost of each type of misclassification, and compute a performance score from the expected cost. 

In machine learning, classifier responses are obtained experimentally and performance scores are computed directly from them, without any assumptions on $P_s$ and $P_n$. Here, we argue that the omitted step of estimating theoretical distributions for $P_s$ and $P_n$ can be useful. In a biometric security example, we fit beta distributions to the responses of two classifiers, one based on logistic regression and one on ANNs, and use them to establish a categorisation into a small number of classes with different extremal behaviours at the ends of the ROC curves.

\end{abstract}

\input{sec1}

\input{sec2}

\input{sec3}

\input{sec4}

\bibliographystyle{plain}
\bibliography{ref}
\end{document}

%% file: sec1.tex
\section{Introduction} 
\label{sec:introduction} 

Machine learning based binary classifiers usually work by processing the input 
to produce a response $r$, most often a real number 
normalised in the interval [0,1], and then comparing $r$ with 
a threshold $t$ and accordingly assigning the input into one 
of the two classes. This is a standard paradigm followed 
even by the newest approaches to the binary classification problem, 
where layers added on top of pretrained deep neural networks such 
as ResNet \cite{Russakovsky:2015} or U-Net \cite{Ronneberger:2015}  
transform the output into a scalar response $r$, which is then compared to a threshold.

The motivation for this paper is the observation of a dichotomy in the practice 
of binary classifier analysis and assessment. In machine learning, where classifiers 
do compute response values, performance measures are computed directly from the responses, 
without fitting to them theoretical distributions $P_s$ and $P_n$ of the probability that the 
response belongs to either class. In other words, performance measures are computed 
from the empirical distributions of $P_s$ and $P_n$. 
In contrast, when the binary classification task is performed by humans the response 
value $r$ is unknown, and since it lacks any obvious physical meaning, there is no 
attempt to estimate it. Instead, {\em detection theory} is employed to estimate theoretical probability 
distributions from the outcome of the experiment, \cite{Macmillan:2004}. Signal detection is 
also employed in the analysis of medical diagnostic 
tests returning black box binary outcomes, \cite{Obuchowski:2018}. 

Here, our goal is to bridge this dichotomy and illustrate the benefits of 
adopting a detection theory approach to the analysis of machine learning binary 
classifiers. Working on a case study from the area of biometric liveness detection, 
we perform the intermediate step of computing theoretical distributions for the 
responses of a machine learning classifier and use them 
to analyse the classifier's behaviour. Specifically, we fit maximum likelihood beta distributions and 
compare the behaviour of two different classifiers, on four datasets of varying difficulty, 
using either cross or within subject validation.

\subsection{Background}

Measuring classifier performance is a challenging problem. 
Various widely used performance measures include  
$$ 
TPR = \displaystyle{\frac{TP}{TP+FN}}, \quad\quad\quad
PPV = \displaystyle{\frac{TP}{TP+FP}}, \quad\quad\quad
 F_1 =  \displaystyle{\frac{2}{\frac{1}{TPR} + \frac{1}{PPV}}} 
$$
where {\em TP, FP} and {\em FN} denote the numbers of true positves, false positives
and false negatives, respectively. They assume a fixed, optimal 
threshold value, while others, such as the {\em Area Under 
the Curve} (AUC) or the Smirnov-Kolmogorov statistic, 
consider a variable threshold and attempt to address the 
trade-off between the two misclassification types, see 
for example \cite{Hand:1997,Pepe:2003,Fawcett:2006}. 
The latter case is further complicated by the fact 
that the misclassification costs of false positives and negatives 
may be application dependent and in some cases cannot 
even be assigned fixed values but are assumed to follow probability 
distributions $P_{fp}$ and $P_{fn}$. 

In \cite{Hand:2009}, it was shown that 
various commonly used performance measures admit a mathematical description as 
the expected misclassification cost under certain assumptions about $P_{fp}, P_{fn}$, 
which in the case of the Smirnov-Kolmogorov statistic are unrealistic convenience 
assumptions, while in the case of AUC depend on  $P_s, P_n$. 
In response to those evident limitations of these measures, the {\em H-measure} was 
proposed which models $P_{fp}, P_{fn}$ as beta distributions, \cite{Hand:2009}. 

Apart from modelling misclassification costs $P_{fp}, P_{fn}$, beta distributions are also 
a natural choice for modelling $P_s, P_n$, as they conveniently have support in 
[0,1] and come in a diversity of shapes. However, they are not used for such purposes 
as widely as normal distributions, perhaps because in many applications 
a total number of four independent parameters is considered large. 
In \cite{Hand:2013}, it was shown that the H-measure can alternatively 
be directly derived from classifier responses, without reference to beta distributed 
misclassification costs. In \cite{Batchelder:2012,Oravecz:2014}, the cultural knowledge of individuals is 
modelled by two beta distributions, but the statistical analysis does not include the construction of 
ROC curves.  In \cite{Steyvers:2014}, beta distributions are used to model probabilistic human 
forecasts and ROC curves are constructed using a hierarchical (group/individual) Bayesian approach. 

In \cite{Gneiting:2018}, beta distributions are used to model the ROC curve itself rather 
than $P_s, P_n$. Their analysis shows superior properties than normal distribution modelling, 
especially when ROC 
curve concavity is required. In \cite{Mossman:2016}, concave ROC curves are constructed 
by modelling $P_s, P_n$ with a pair of {\em dual} beta distributions with two free variables. 
Here, as concavity is a constraint we want to avoid, we do not restrict the parameter range of 
the beta curves. Indeed, in practice, including the classifiers analysed in this paper, it is quite 
common for the empirical ROC curve of a machine learning classifier to be under the diagonal near 
0 or 1. 

Finally, in an approach that is most similar to ours, the assessment of 
medical diagnostic tests with continuous random variables as outcome often uses 
theoretical distributions estimated from the test's scalar responses, \cite{Cai:2002}. 
However, since medical data are usually sparse, the fitted distributions are very 
simple, usually normal and quite often with equal standard deviations, \cite{Cai:2004}.

\subsection{Contributions and limitations} 

Compared to the commonly used machine learning performance measures, 
our approach gives novel insights to the behaviour of a classifier, which  
is not straightforward to gain directly from the empirical response distributions. 
In particular: 
\begin{enumerate} 
\item By using continuous theoretical ROC curves we can approach naturally 
questions related to ROC curve derivatives. In particular, what are the rates of true 
positives to false negatives as $t$ approaches the extreme values 0 or 1? 
\item Properties of the theoretical distributions can be used to categorise classifier-dataset 
combinations into a small number of classes. In our case study, we note that shape variations 
between U-shaped and J-shaped distributions indicate qualitative differences in behaviour. 
\end{enumerate} 

The main limitation of the proposed approach is that our choice of family of theoretical 
distributions is to some extent arbitrary and other natural choices, such as mixtures 
of Gaussians, would lead to different analyses. Moreover, the method of fitting theoretical 
distributions to the data, here maximum likelihood, can also influence the results. That means 
that the proposed method does not give a classifier performance measure, but rather 
a tool for analysing classifier behaviour. 

Indeed, a desirable characteristic of performance measures is that they should be simple 
enough to be reported as single numbers for each classifier-dataset combination, 
\cite{Anagnostopoulos:2012}, which is not the case in our approach. However, as noted in 
\cite{Hand:2014}, beyond the problem of objective comparisons between classifiers, there is 
the different problem of analysing their behaviour during development where, for example, the use of 
H-measures with researcher defined beta distributions may be justified. Going a step further, 
we note that during development, where intuitiveness and insight can be as useful as 
objectivity, computing single numbers, such as the H-measure, should not be considered 
necessary requirement. 


%% file: sec2.tex
\section{Fitting beta distributions to binary classifier responses} 
\label{sec:betas}

\begin{figure}[t]
	\centering
		\includegraphics[width=0.19\columnwidth]{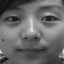} \hfill 
		\includegraphics[width=0.19\columnwidth]{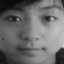} \hfill 
		\includegraphics[width=0.19\columnwidth]{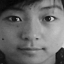} \hfill 
		\includegraphics[width=0.19\columnwidth]{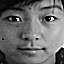} \hfill 
		\includegraphics[width=0.19\columnwidth]{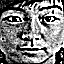} 
		\caption{Test set images. From left to right: Client, impostes and imposters  sharpened by 1.0, 5.0 and 50.0, respectively.}
		\label{fig:face}
\end{figure} 

Our binary classifiers are trained for {\em face liveness detection} from still images, 
that is, classifying face images into the {\em client} and {\em imposter} classes. 
Clients are images of real human faces captured by the camera of a face recognition 
system, while imposters are images of photographs of human faces displayed in front 
of the system's camera. 

The first classifier is sparse logistic regression (SLR) on differences of 
Gaussians of the images, see \cite{Tan:2010}, and the second is an Artificial 
Neural Network (ANN) with a single hidden layer of 10 nodes trained on raw 
images. We used the $64\times 64$ greyscale images of the NUAA database in 
\cite{Tan:2010}, and trained classifiers with either {\em cross-subject} or 
{\em within-subject} protocols. Using Matlab's {\em imsharpen} function, 
we processed the imposter images of the test set with three different amounts of 
sharpening, creating a family of four different test sets of increasing difficulty, 
see Figure~\ref{fig:face}. This step was based on the observation that as imposter
images generally lack high frequency information, attacks with sharpened imposter 
images should be more challenging to detect, \cite{Omar:2016a}. 

In total, using two classifiers, two training protocols and four datasets, we 
created a $2 \times 2 \times 4$ space to observe variations in the shape of the 
theoretical distributions of the responses. In all cases, we fitted two maximum 
likelihood beta distributions, one on responses on imposters and one on clients, 
using Matlab's {\em betafit } function. The computational time for fitting 
a beta distribution on 1000 responses was approximately16ms on a macOS 
with 2.3GHz i5 CPU, 8GB 2133MHz LPDDR3 RAM and an Iris Plus Graphics 640 1536MB. 
The low computational cost means that the method is fast enough to be used repeatedly  
for systematic parameter optimisation, or in large 
multi-parameter ablation studies for gaining insights into the classifier's performance 
on complex tasks.  


\subsection{Results} 
\label{sec:results}

\begin{figure*}[t]
	\centering
		\includegraphics[width=0.19\textwidth]{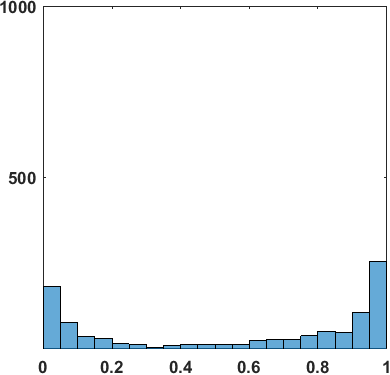} \hfill  
		\includegraphics[width=0.19\textwidth]{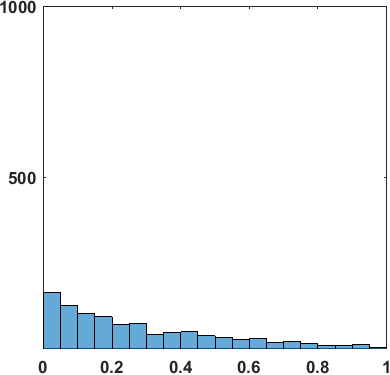} \hfill  
		\includegraphics[width=0.19\textwidth]{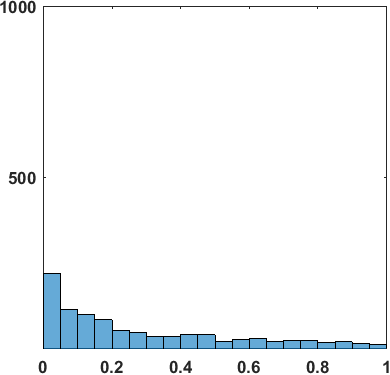} \hfill  
		\includegraphics[width=0.19\textwidth]{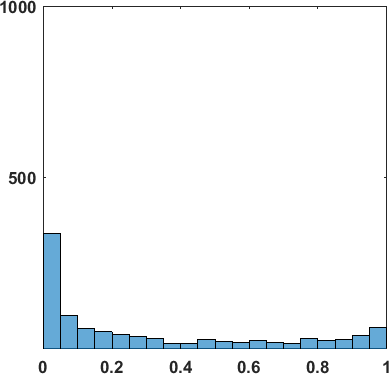} \hfill  
		\includegraphics[width=0.19\textwidth]{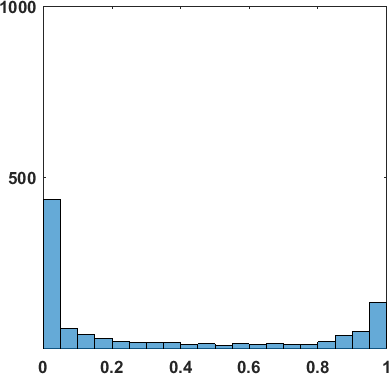} 
		\includegraphics[width=0.19\textwidth]{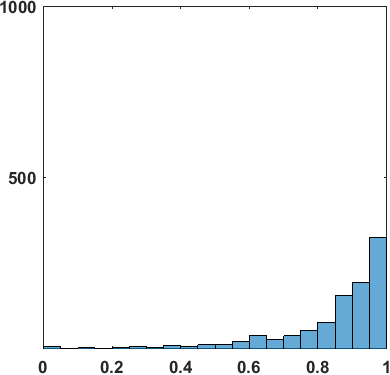} \hfill  
		\includegraphics[width=0.19\textwidth]{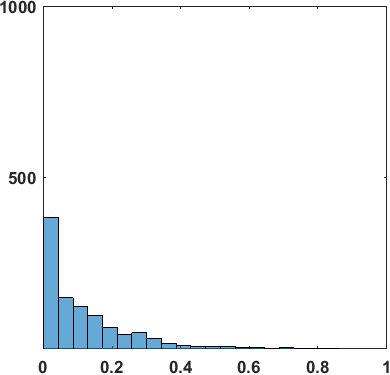} \hfill  
		\includegraphics[width=0.19\textwidth]{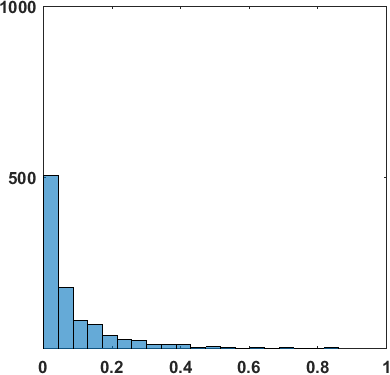} \hfill  
		\includegraphics[width=0.19\textwidth]{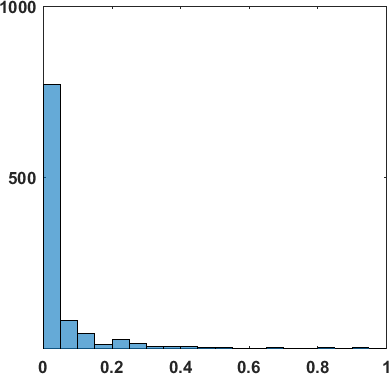} \hfill  
		\includegraphics[width=0.19\textwidth]{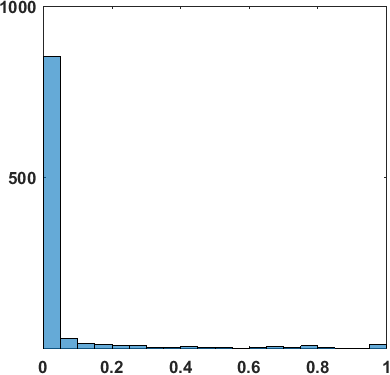}  
		\includegraphics[width=0.19\textwidth]{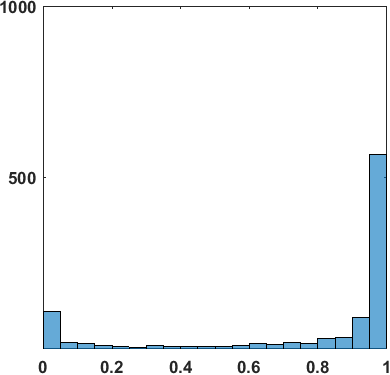} \hfill 
		\includegraphics[width=0.19\textwidth]{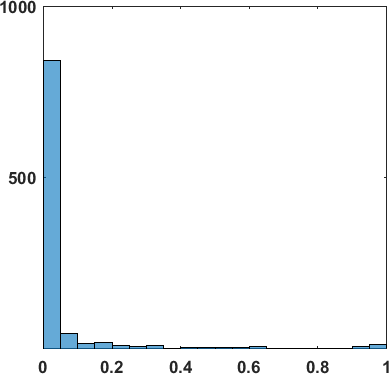} \hfill 
		\includegraphics[width=0.19\textwidth]{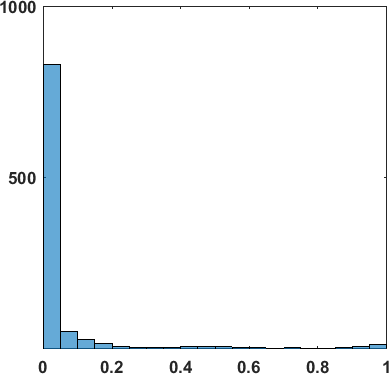} \hfill 
		\includegraphics[width=0.19\textwidth]{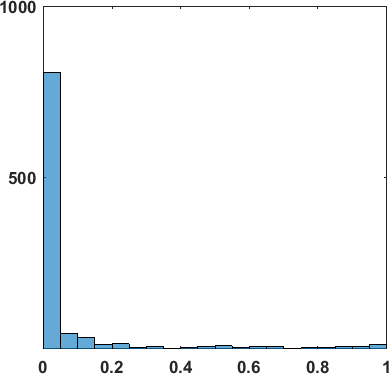} \hfill 
		\includegraphics[width=0.19\textwidth]{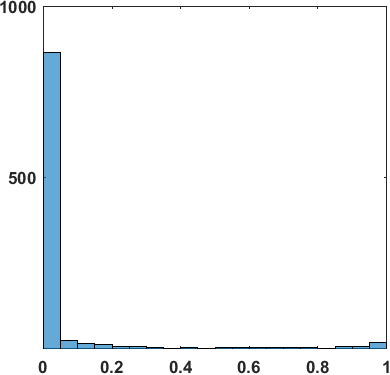} 
		\includegraphics[width=0.19\textwidth]{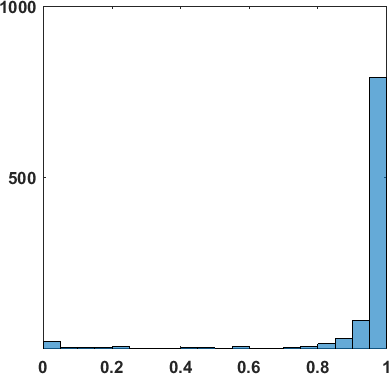} \hfill  
		\includegraphics[width=0.19\textwidth]{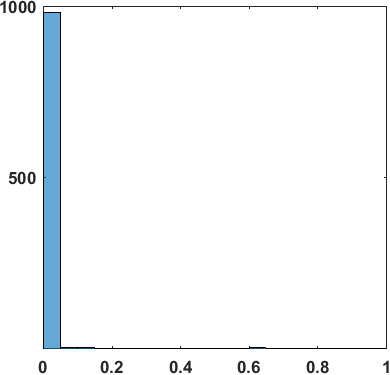} \hfill  
		\includegraphics[width=0.19\textwidth]{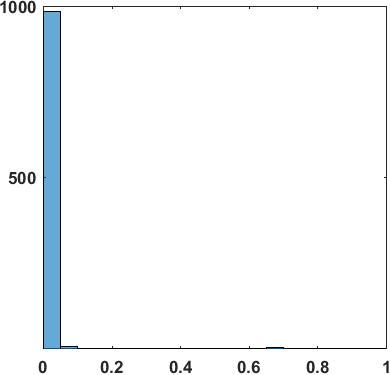} \hfill  
		\includegraphics[width=0.19\textwidth]{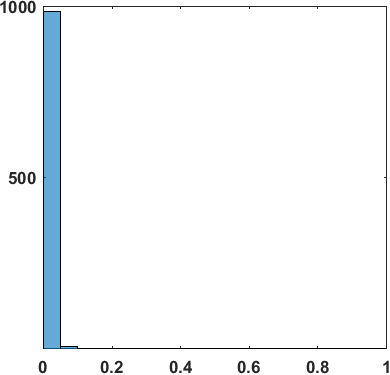} \hfill  
		\includegraphics[width=0.19\textwidth]{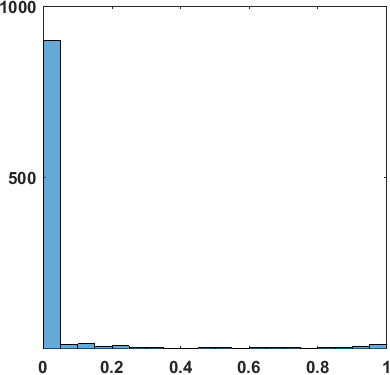} 
	\caption{{\bf Left to Right:} Twenty-bin histograms of clients, imposters and imposters sharpened by 1.0, 5.0, and 50.0, respectively. {\bf Top to Bottom:} Cross-subject SLR, within-subject SLR, cross-subject ANN and within-subject ANN.}
	\label{fig:results2}
\end{figure*}

\begin{figure*}[t]
	\centering
		\includegraphics[width=0.24\textwidth]{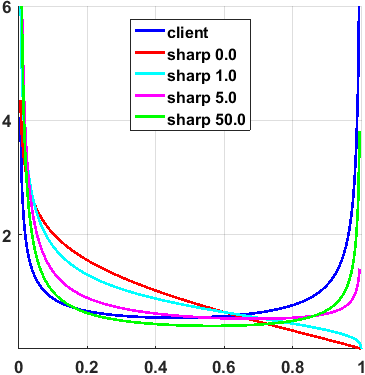} \hfill 
		\includegraphics[width=0.24\textwidth]{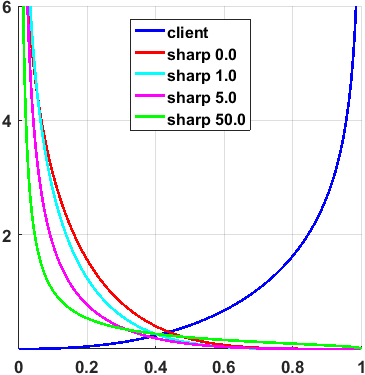} \hfill
		\includegraphics[width=0.24\textwidth]{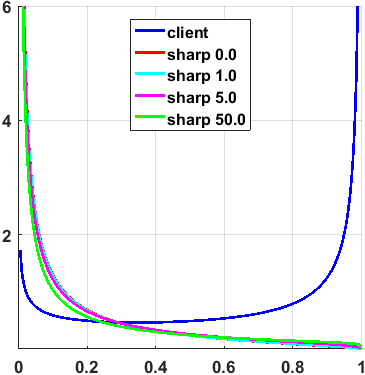} \hfill
		\includegraphics[width=0.24\textwidth]{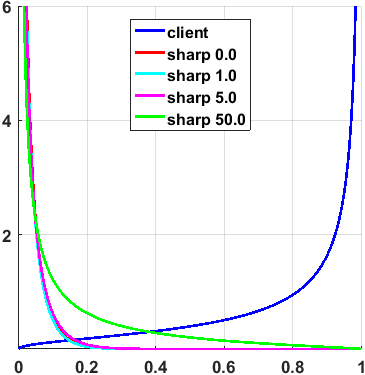} 
		\caption{Fitted beta distributions. {\bf Left to right:} Cross-subject SLR, within-subject SLR, cross-subject ANN and within-subject ANN.}
		\label{fig:results3}
\end{figure*}

Figure~\ref{fig:results2} visualizes the empirical distributions of the responses of the four 
classifiers as twenty-bin histograms. The first column shows the responses on client images, 
which are the same over all four test sets, while the other columns show the responses on 
imposters for each test set. We notice that the skewness may vary considerably 
between histograms, further justifying the use of beta rather than normal 
distributions. Figure~\ref{fig:results3} shows the plots and Table~\ref{table:beta} shows 
the values of the $\alpha, \beta$ parameters of the fitted distributions.

\begin{table}[t]
	\caption{Maximum likelihood estimates of the $\alpha,\beta$ parameters of 
	the beta distributions.} 
	\centering 
	\begin{tabular}{|c|c|c|c|c|} 
	\hline 
	& \multicolumn{2}{|c}{SLR} &  \multicolumn{2}{|c|}{ANN} \\ \hline 
 	& cross & within & cross & within \\ \hline 
client	& $0.47,0.36$ & $3.27,0.67$ & $0.61,0.27$ & $1.47,0.29$	\\ 
imp0	  & $0.77,1.91$ & $0.71,5.04$ & $0.18,1.66$ & $0.24,17.5$	\\ 
imp1	  & $0.59,1.36$ & $0.57,5.39$ & $0.18,1.63$ & $0.23,17.8$	\\ 
imp5	  & $0.34,0.70$ & $0.30,4.26$ & $0.17,1.38$ & $0.21,14.2$	\\ 
imp50   & $0.22,0.39$ & $0.13,1.39$ & $0.14,1.12$ & $0.17,1.79$	\\ \hline 
	\end{tabular} 
	\label{table:beta} 
\end{table}

%% file: sec3.tex
\section{Discussion} 
\label{sec:discussion}

\subsection{Shapes of the beta distributions}

Excluding the singular cases $\alpha = 1, \beta = 1$ and 
$\alpha = \beta$, the shape of the beta distribution is {\em bell-shaped} if 
$\alpha,\beta > 1$, {\em U-shaped} if $\alpha,\beta < 1$ and {\em J-shaped} or 
reverse J-shaped if 
$\alpha < 1 < \beta$ or $\beta < 1 < \alpha$, respectively, see \cite{owen2008}. 
Depending on whether J-shaped and reverse J-shaped shapes are considered 
separately, we have a total of 9 or 16 categories for the shapes of the 
distributions of the client and imposter responses.

In all our experiments the obtained distributions were either
{\em J-shaped} or {\em U-shaped}. The client distributions were 
{\em U-shaped} in both cross-subject validation cases and 
{\em J-shaped} in both within-subject, reflecting the more 
challenging nature of cross-subject validation. Indeed, 
when the client distribution is U-shaped, the rate of reduction of 
the false negatives decreases as the threshold approaches 0, while 
a U-shaped imposter distribution means a decreasing rate of reduction 
of the false positives as the threshold approaches 1. 

Regarding the comparison between imposter and sharpened 
imposter distributions, we notice that $\alpha$ is always less 
than 1 and decreases with sharpening, while $\beta$ in most 
cases is greater than 1, giving J-shaped imposter distributions. 
The notable exception is in the cross-subject SLR case for the 
two larger amounts of sharpening where $\beta$ is less than 1 
and the distribution becomes U-shaped, indicating the need to 
operate on strict thresholds to reduce significantly the number 
of false positives. In all other cases, the sharpened imposter 
distributions remain J-shaped, however, their tail increases 
with sharpening, bringing them closer to a U-shaped distribution. 
The larger tails are also noticeable in the empirical distributions in 
Figure~\ref{fig:results3}.


\subsection{Extremal properties of the ROC curves} 

A decision on whether to include a weak classifier in an ensemble may depend on 
its behaviour at thresholds near 0 or 1, in particular, on whether its ROC curve 
is above or below the diagonal line of no-discrimination near 0 or 1. Assuming 
a continuously differentiable ROC, that depends on its right semi-derivative at 
0 and its left semi-derivative at 1. 

Let $(\alpha_1, \beta_1)$ and $(\alpha_2, \beta_2)$ be the beta parameters 
of clients and imposters, respectively. Up to a constant, the respective 
cumulative distributions $F_{c}(x)$ and $F_{i}(x)$ are 
$$
F_{c}(x) = 
\int_{0}^{x} t^{\alpha_1 - 1}(1-t)^{\beta_1 - 1} dt \quad\quad\quad\quad F_{i}(x) = 
\int_{0}^{x} t^{\alpha_2 - 1}(1-t)^{\beta_2 - 1} dt
$$
and the derivative of the ROC curve 
$R(x) = (F_{i}(x), F_{c}(x))$ in (0,1) is 
$$ 
\frac{dF_{i}(x)/dt}{dF_{c}(x)/dt} = 
\displaystyle{\frac{x^{\alpha_2 - 1}(1-x)^{\beta_2 - 1}}{x^{\alpha_1 - 1}(1-x)^{\beta_1 - 1}}} = 
x^{\alpha_2 - \alpha_1} (1-x)^{\beta_2 - \beta_1}
$$
The right semi-derivative at 0 is 
$\displaystyle{\lim_{x\rightarrow 0^+}} x^{\alpha_2 - \alpha_1}$, 
i.e. 0 when $\alpha_1 < \alpha_2$ and $\infty$ when $\alpha_1 > \alpha_2$. 
Similarly, the left semi-derivative at 1 is 
$\displaystyle{\lim_{x\rightarrow 1^-}} (1-x)^{\beta_2 - \beta_1} = 
\displaystyle{\lim_{x\rightarrow 0^+}} x^{\beta_2 - \beta_1}$, 
i.e. 0 when $\beta_1 < \beta_2$ and $\infty$ when $\beta_1 > \beta_2$.
From the continuity of the derivative in (0,1), we have that 
the ROC curve is above the diagonal near 0 when the semi-derivative 
at 0 is $\infty$, i.e. when $\alpha_1 < \alpha_2$. Similarly, near 1 the 
ROC curve is above the diagonal when the semi-derivative at 1 is 0, 
i.e. when $\beta_1 > \beta_2$. 

From Table~\ref{table:beta} we notice that in all cases, apart from 
the two exceptions of cross-validated SLR classifiers with  
sharpening parameters 0 and 1, we have $\alpha_1 > \alpha_2$, 
the semi-derivative at 0 is $\infty$ and thus, the ROC curve 
stays above the diagonal near 0. That indicates that weak ANN based 
classifiers operating at thresholds close to 0  can be included 
in an ensemble. The situation is different with the SLR 
classifiers, the low performance of which on unseen faces at thresholds 
near 0 makes them unsuitable. A similar comment on the limitations 
of SLR based classifiers at thresholds near 0 was made in \cite{Tan:2010}, 
but it was based on visual inspection of the empirical ROC curves. 

We also notice that the effect of sharpening is a decrease of 
the value of $\alpha_2$, resulting to an increase of the value of 
$\alpha_2 - \alpha_1$, which means that the ROC curves rise at 
a slower pace near 0. That means that against weak classifiers operating at thresholds 
close to 0, sharpening the imposter images can be an effective 
attacking technique. 

Regarding the behaviour of the theoretical ROC curves near 1, we 
notice that in all cases the left semi-derivative is 0, 
meaning that the ROC curves stay above the diagonal, indicating 
suitability for inclusion in ensembles of weak classifiers 
operating at thresholds close to 1. We also notice that the 
difference $\beta_2 - \beta_1$ decreases for the two larger 
amounts of sharpening, meaning that the ROC curves level-off at 
slower pace, again indicating the effectiveness of the sharpening attack. 

\begin{figure*}[t]
	\centering
		\includegraphics[width=0.24\textwidth]{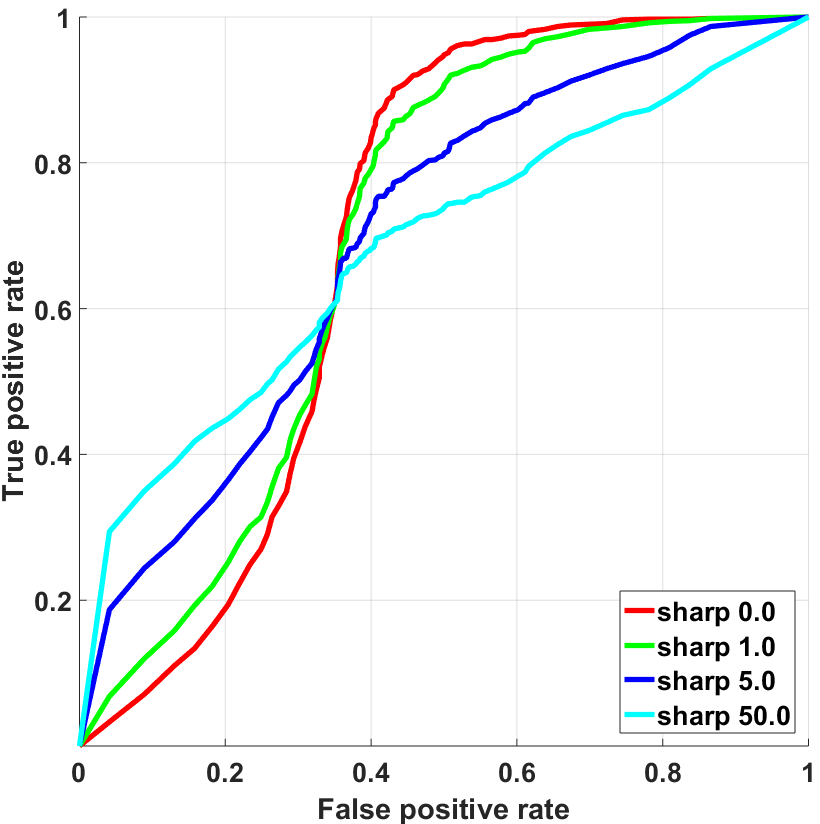} \hfill 
		\includegraphics[width=0.24\textwidth]{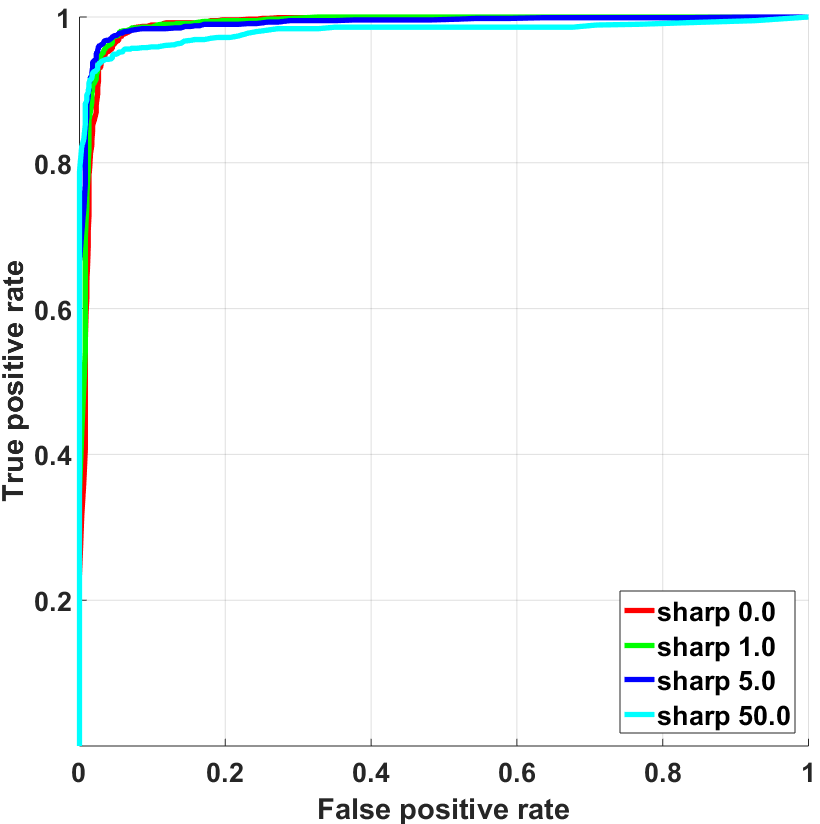} \hfill  
		\includegraphics[width=0.24\textwidth]{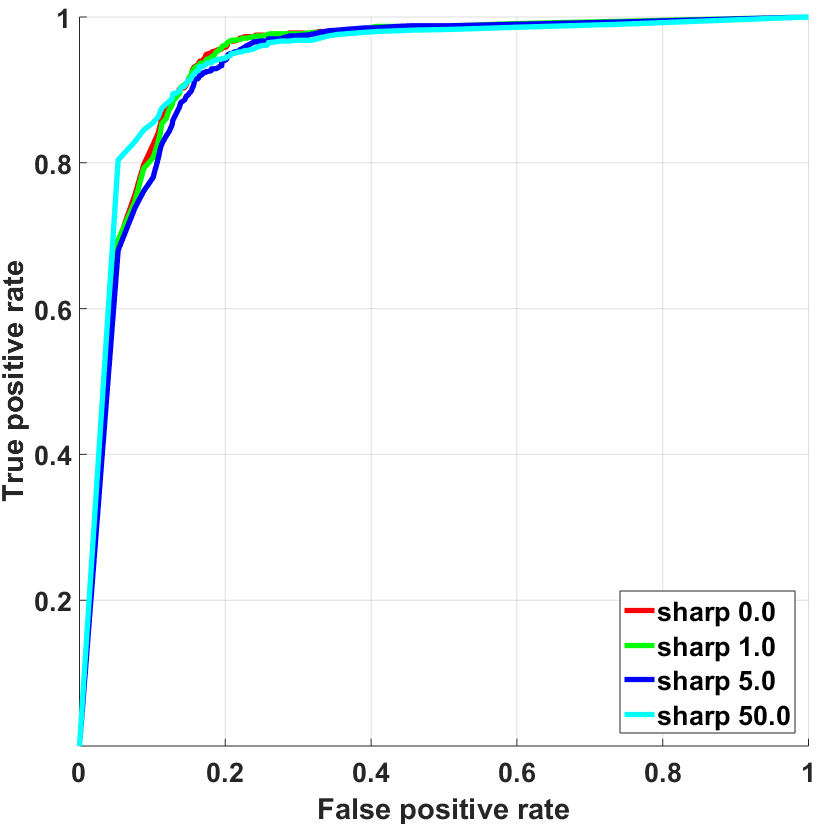} \hfill  
		\includegraphics[width=0.24\textwidth]{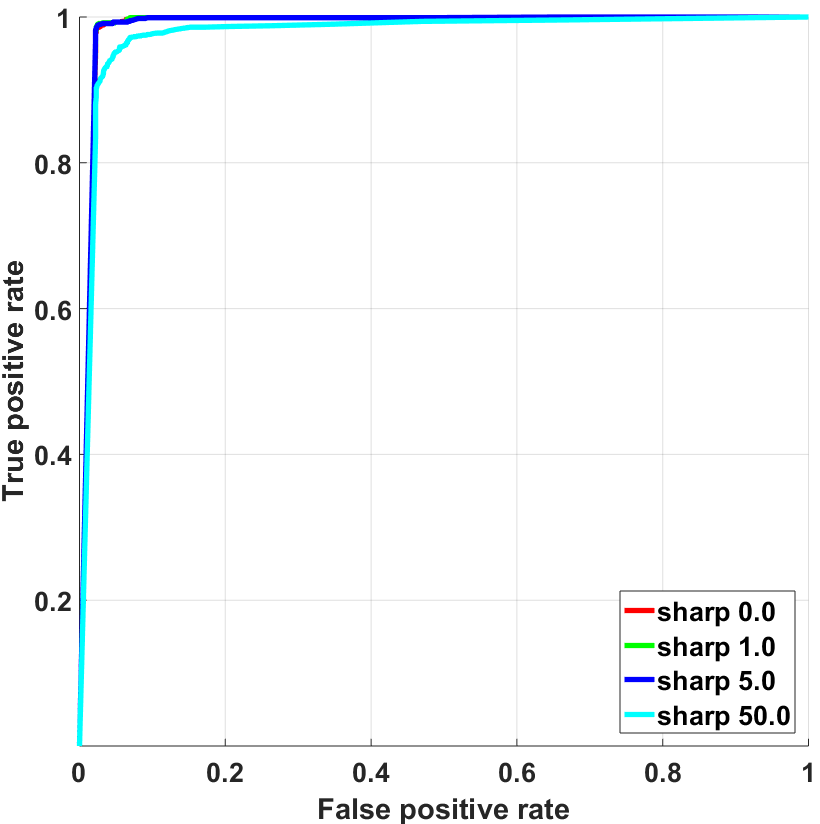}  
		\includegraphics[width=0.24\textwidth]{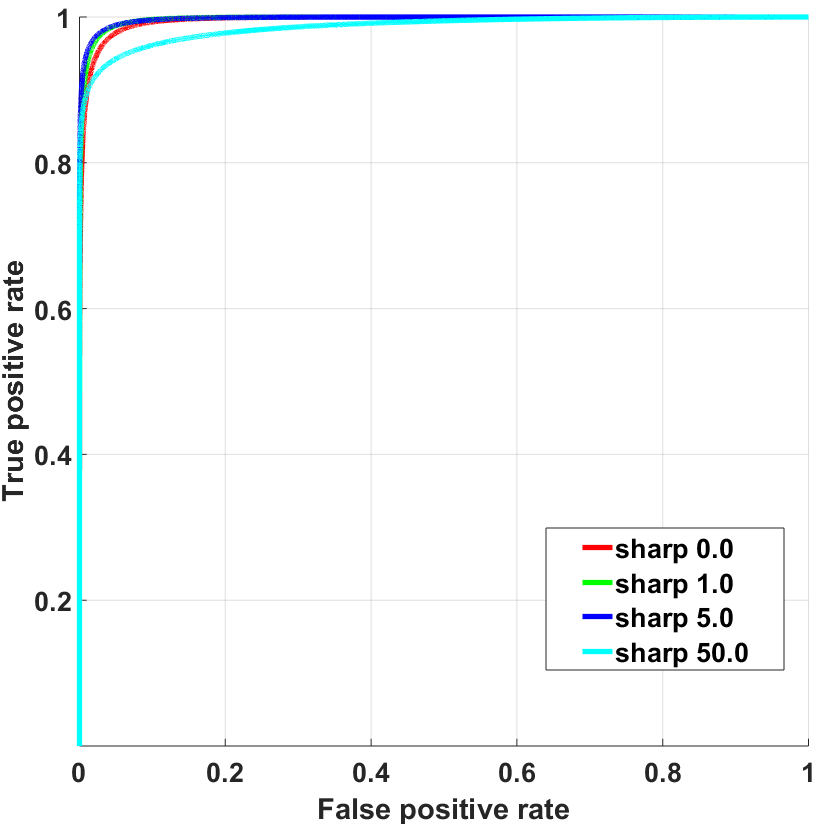} \hfill 
		\includegraphics[width=0.24\textwidth]{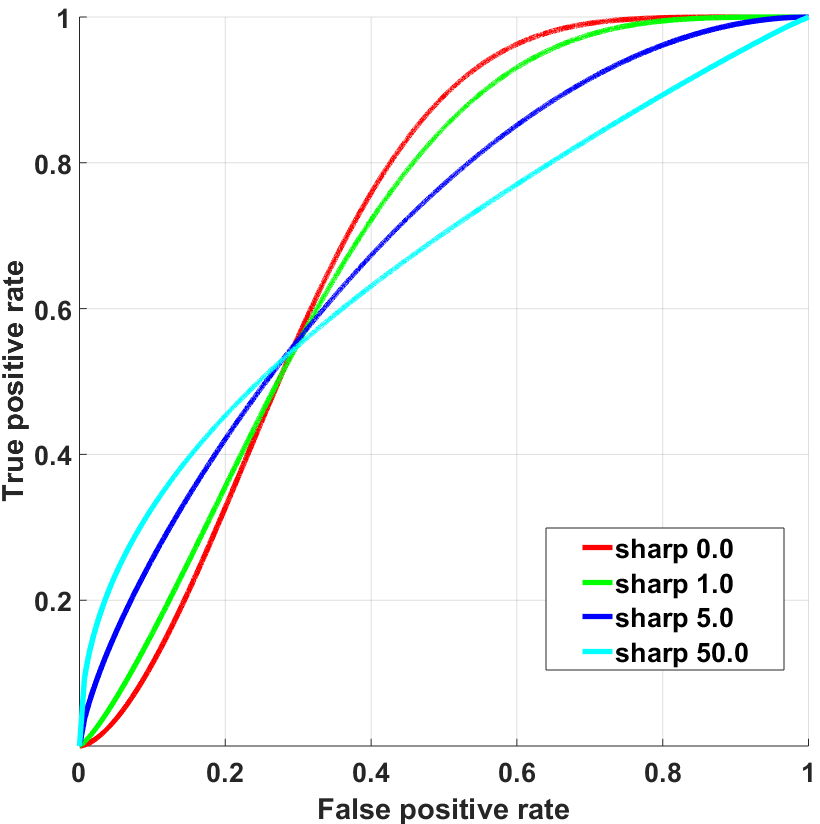} \hfill  
		\includegraphics[width=0.24\textwidth]{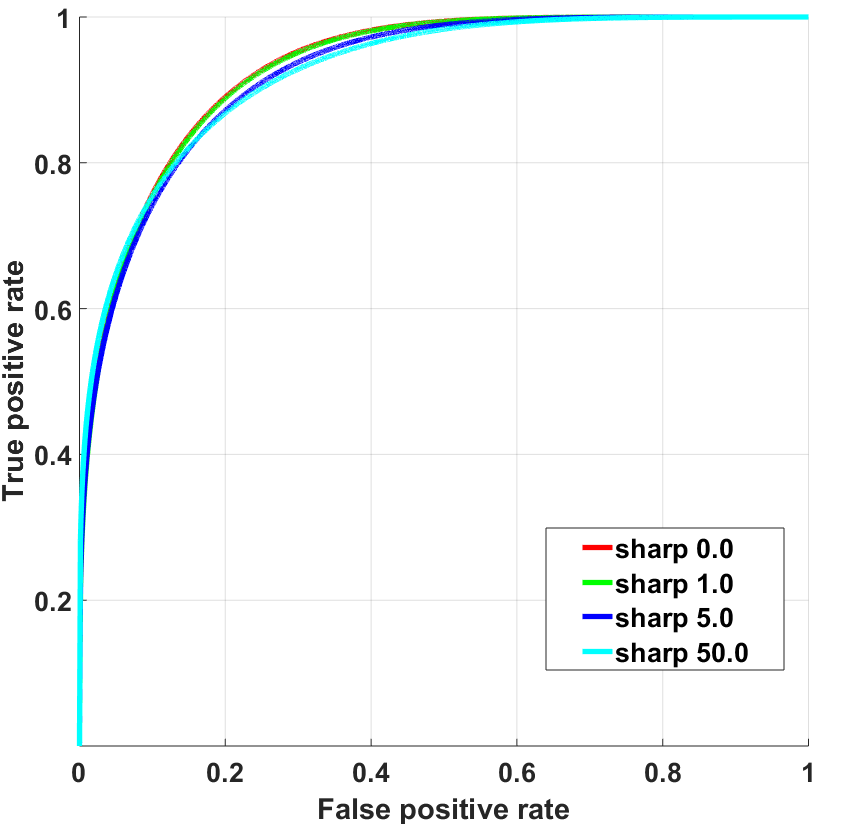} \hfill  
		\includegraphics[width=0.24\textwidth]{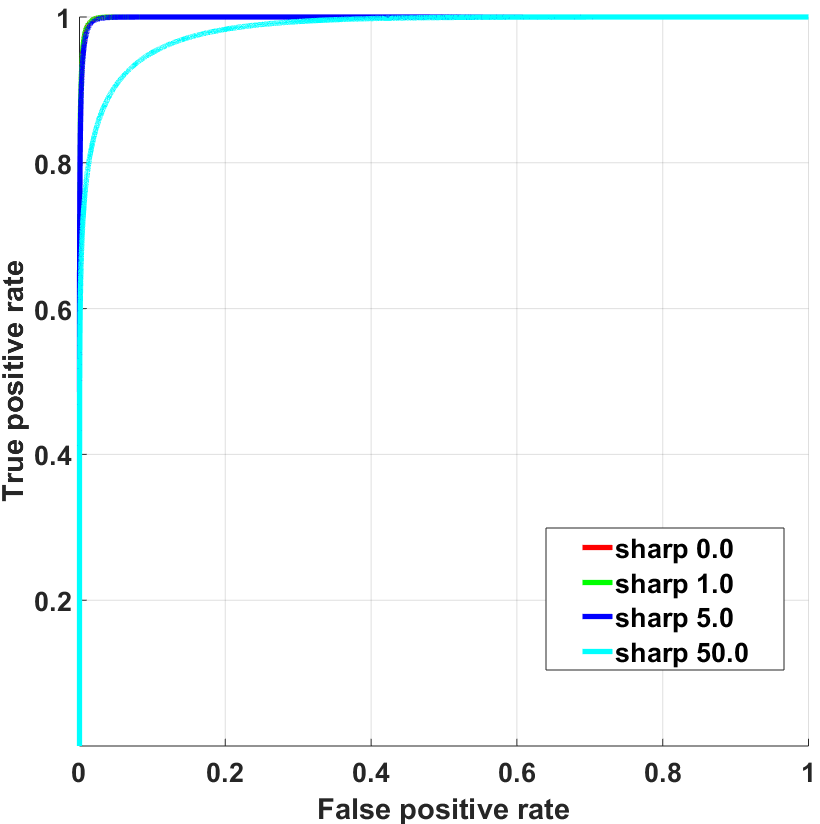} 
		\caption{Empirical (top) and theoretical (bottom) ROC curves. 
		{\bf Left to right:} SLR cross-subject, SLR within-subject, ANN cross-subject, ANN within-subject.}
		\label{fig:results1}
\end{figure*}

Figure~\ref{fig:results1} shows the empirical 
and theoretical ROC curves. Here, positive 
tags correspond to classification as imposter. 

%% file: sec4.tex
\section{Conclusion} 
\label{sec:conclusion} 

We propose the use of theoretical ROC curves 
for analysing the behaviour of machine learning binary classifiers. 
While our approach does not provide a new objective classifier performance 
measure, we demonstrated its usefulness as an analytical tool 
facilitating classifier development. 

The proposed continuous approach was used to get insights 
that can be easily lost in a discrete setting. We studied in a natural way 
classifier behaviour near the two ends of the threshold range 
and showed that beta distributions in particular have the 
expressive power to model that behaviour. We note that 
while any computation one can do with continuous ROC 
curves can in principle be performed directly on the discrete set 
of responses, to the best of our knowledge, the question of 
performance at the extrema of the threshold range 
has not been previously addressed in such a systematic way. 
Finally, we showed that the use of theoretical ROC curves based on beta distributions 
can lead to a natural categorization of classifier behaviour into a few classes, 
an approach to the analysis of classifiers which, to the best of our knowledge, 
has be overlooked as far as machine learning classifiers are concerned. 

In the future, we would like to further study theoretical ROC curves based 
on beta distributions and discover properties that could be relevant to the behaviour 
of machine learning binary classifiers. The use of very large synthetic 
datasets, allowing the dense sampling of the parameter space of the synthetic 
data generator, would greatly facilitate such a study.